\documentclass[preprint,12pt]{elsarticle}

\usepackage{amssymb}
\usepackage{amsmath}
\usepackage{amsfonts}
\usepackage{graphicx}
\usepackage{multicol}
\usepackage{subfigure}

\journal{Decision Support Systems}

\begin{document}

\begin{frontmatter}



\title{Joint learning of interpretation and distillation}


\author{Jinchao Huang\fnref{label1}}
\ead{hjc2015@sjtu.edu.cn}

\author{Guofu Li\fnref{label2}}
\ead{li.guofu.l@gmail.com}

\author{Zhicong Yan\fnref{label1}}
\ead{zhicongy@sjtu.edu.cn}

\author{Fucai Luo\fnref{label3}}
\ead{519594474@qq.com}

\author{Shenghong Li\corref{cor1}\fnref{label1}}
\ead{shli@sjtu.edu.cn}

\cortext[cor1]{Corresponding author.}
\fntext[label1]{School of Cyber Security, School of Electronic Information and Electrical Engineering, Shanghai Jiao Tong University, Shanghai 200240, China.}
\fntext[label2]{AI Lab of Ping An Asset Management Co., Ltd, No. 1333 Lujiazui Huan Road, Shanghai, China.}
\fntext[label3]{State Grid Fujian Electric Power Company, No.257 Wusi Road,Fuzhou City, Fujian Province, P.R. China}

\begin{abstract}
The extra trust brought by the model interpretation has made it an indispensable part of machine learning systems.
But to explain a distilled model's prediction, one may either work with the student model itself, or turn to its teacher model.
This leads to a more fundamental question: if a distilled model should give a similar prediction for a similar reason as its teacher model on the same input?
This question becomes even more crucial when the two models have dramatically different structure, taking GBDT2NN for example.
This paper conducts an empirical study on the new approach to explaining each prediction of GBDT2NN, and how imitating the explanation can further improve the distillation process as an auxiliary learning task. Experiments on several benchmarks show that the proposed methods achieve better performance on both explanations and predictions.
\end{abstract}

%


\end{frontmatter}


\section{Introduction}
\label{lable_Intro}
When Gradient Boosting Machines (GBMs) show consistently satisfactory performance on noisy and heterogeneous data \cite{caruana2006empirical, wu2010adapting, zhang2015gradient, li2005scalable, wu2019feature}, it has two major weak points: 1) its ineffectiveness on sparse categorical features, and 2) its lack of proper mechanism to make online update. Neural network(NN), on the other hand, shows better capabilities on these two points, despite its weakness on dense numerical features.

Therefore, efforts have been made to combine the advantages of GBM and NN.
A straight-forward way to combine the two is to concatenate them directly \cite{he2014practical}.
Soon after, Ling et al. \cite{ling2017model} proposed to use the predicted values of NN to initialize the gradient boosting decision tree(GBDT), and then the results learned by GBM are used to fit the residual of NN.
It is also pointed that the boosted deep neural network(DNN) is the best among all of these combining methods \cite{ling2017model}.
Nevertheless, these methods fail to fullfil the online updating requirements.
Thus, researches have been made to convert the decision tree to NN model. For instance, some works \cite{banerjee1997initializing, humbird2019deep} try to use the results as well as structure learned by decision tree to initialize NN.
There is also proposed the method that retained the frame of random forest, and replaced the decision tree with NN \cite{biau2016neural}. Some studies \cite{richmond2015relating} explored the relationships between cascaded random forests and convolutional neural network(CNN). When the number of decision trees is large, such conversion solutions have to construct a very wide NN to represent, making these approaches extremely inefficient or even impractical.

To solve this problem, Ke et al. \cite{ke2019deepgbm} put forward a new framework, namely GBDT2NN, which makes use of the {\em knowledge distillation} to approximate the function of the tree structure by an NN.
There are mainly three innovations of GBDT2NN when compared with previous approaches: 1) It only takes the features that decision tree used as the inputs, 2) Then it adds an embedding layer between the leaf index and the output to reduce the dimension of structure distillation, 3) Finally it proposes the {\em Tree Grouping} strategy to reduce the number of distilled NNs.
This by far has been one of the best options to incorporate the two different architectures.

However, learning to give the same answer is just one side of knowledge distillation. We argue that besides ``learn how'' task, an intelligent agent should also care about ``learn why'' task.
That is, distillation should cover both the prediction output and prediction explanation of the teacher model.
Model interpretability, as a mean to boost end-users' trust and to help developers to understand and diagnose the model, makes up for an indispensable part of machine learning systems \cite{du2019techniques, chatterjee2019explaining, al2013people, lee2010using}.
There have been various studies on interpreting the behaviors of both GBMs and NNs though, very few studies have been spotted on how NN model derived via GBM might be interpreted, as well as how this auxiliary task can help to further improve the model distillation. Therefore, this paper proposes to take use of the interpretations of the teacher model to get the local explanation of distillation model.
Following the previous researches, we suggest that this problem can be better-handled by a two-stage framework: building an internal repetition vector of the teacher model, which then helps to derive the interpretation vector for the student model. Within this framework, we propose two types of methods: the {\em independent method} trains a new model to provide explanations for the distillation method, while the {\em joint method} combines the distillation model and explanation model together by a two-folder loss function. To verify these ideas, we conduct several experiments on  publicly available datasets, which demonstrate that the proposed methods can achieve good performance on both explanations and predictions.


The rest of this paper is organized as follows. In Section \ref{sec:related-works}, we give a brief introduction on existing approaches on prediction explanation.
In section \ref{sec:GBDT2NN}, we introduce the GBDT2NN model. And the proposed interpretable methods for the distilled model is described in section \ref{sec:distill}. Descriptions of the datasets and the experiment results are shown in section \ref{sec:exp}. Section \ref{sec:conclusion} concludes the paper and discusses some guidelines for the future work.

\section{Related Works}
\label{sec:related-works}

\subsection{Interpreting the tree-based model}

As the base predictors, the decision trees prefer to use more leaf nodes to replace the subtrees, and the sparsity characteristics prompt the model to use fewer features to obtain the predictions. This property helps people to know what features drive the model's prediction.
But it is difficult to intuitively understand the tree-based ensemble models, such as XGBoost \cite{chen2016xgboost}, CatBoost \cite{prokhorenkova2018catboost}, lightGBM \cite{ke2017lightgbm} and some other complicated models.

The interpretability methods\cite{eiras2019scalable, vandewiele2016genesim, chebrolu2005feature, auret2011empirical} of ensemble trees can be divided into two categories: {\em global interpretability}, and {\em local interpretability}.

To summarize, interpreting tree-based models is relatively straight-forward. There are two general ways to calculate the {\em global feature importance} values.
\begin{enumerate}
    \item Replace the original model with a simple one (i.e., {\em model extraction}), given that the simple model can approximate its teacher model well enough. Then the statistical characteristics of the original model could be reflected in the simple model.
    \item Measure the contributions of features mainly in some primary forms, which may further be classified into three types:
    \begin{enumerate}
        \item A classic approach is to compute the gain at each split node, which contributes to the importance value of the feature that this node represents.
        \item The second approach is to calculate the changes in the model's error caused by {\em permuting} the values of a feature. Since if a feature is important, its permutation should create a large increase in the model's error.
        \item The third approach is to {\em count} the feature splits, which means how many times a feature is used to split at the internal nodes.
    \end{enumerate}
\end{enumerate}

Local interpretation helps to understand the model behaviour on each of the individual input.
Though some {\em model agnostic} local interpretation methods can be applied to tree-based model, but they are significantly slower than tree-specific methods and have sampling variability.
By far, we consider two widely-applied tree-specific methods: Sabbas \cite{saabas2014interpreting} and SHAP \cite{lundberg2018consistent}.
\begin{itemize}
    \item The Sabbas method measured the change in the model's expected output, and the differences between these expectations on each internal node are treated as the attributes of splits features. So, the Sabbas is a structure-based method.
    \item The SHAP method introduced the concept of SHAP values, then presented Tree SHAP to estimate SHAP values of tree ensembles.
\end{itemize}

\subsection{Interpreting the neural nets}

Modern deep learning models create hierarchies of representation from raw data via layers of transformations, making it an essential step to understand the intermediate layers of NNs in terms of model interpretations in many of the approaches.

Some literature proposed to use the information learned from raw data by NNs as the explanations of models. These methods are mostly designed for some certain scenarios and networks.
Zhang et al.\cite{zhang2018interpretable} added different losses at the higher convolutional layer to know the information learned by the high layers of CNN.
Besides the loss functions, researchers also adopted the attention mechanism, which is often used in sequential models, to grasp features, and then converted the grasped information to a form that can be easily understood by users \cite{xu2015show, bahdanau2014neural}.
Specifically, Xu et al.\cite{xu2015show} attempted to grasp the picture features at a lower convolutional layer during the encoding process, then used Long Short-Term Memory(LSTM) to transform the grasped information into a textual description.
Bahdanau et al. \cite{bahdanau2014neural} applied attention mechanism to the field of machine translation, and treated the corresponding relations between translation results and the original sentence as the interpretations.

Some methods tried to analyze the trained structures and the parameters of NNs to obtain explanations \cite{liu2019representation}. In order to better understand the entire model framework, a lot of works \cite{karpathy2015visualizing, kadar2017representation} focused on capturing the information learned by the intermedia layers of NNs. Interpreting methods of CNN usually use the {\em Activation maximization} (AM) framework to capture the learned information at a certain layer of the preferred inputs,
while recurrent neural network(RNN) related methods often use the language modeling techniques to analyze the neurons.

For the local explanations of individual samples, one way is the investigation of deep representations in intermediate layers as described in global methods, and some others are to return the NNs to a ``white-box'', which is composed of the linearly weighted additions of important features as follows.
Some works directly calculated the gradient of a particular output related to the input by back-propagation to obtain the influence of features \cite{ancona2017towards, bach2015on}.
The perturbation-based methods \cite{fong2017interpretable} sequentially perturbed the input to get the contribution values of each feature. Some studies \cite{du2018towards, du2019attribution} also utilized the representations of the intermediate layers to perform attribution.

On the whole, the first kind of methods are only applicable to certain sceneries, which are not suitable to the GBDT2NN model. The latter methods take use of the parameters and the derivation conduction to explain the NNs, so they are only applicable to the NNs of the end-to-end forms. All of them do not apply to GBDT2NN.

\section{Preliminaries}
\label{sec:GBDT2NN}

For better understanding, we introduce the GBDT2NN model first in this section.

\begin{figure}[htbp]
\centering
\includegraphics[width=0.8\textwidth]{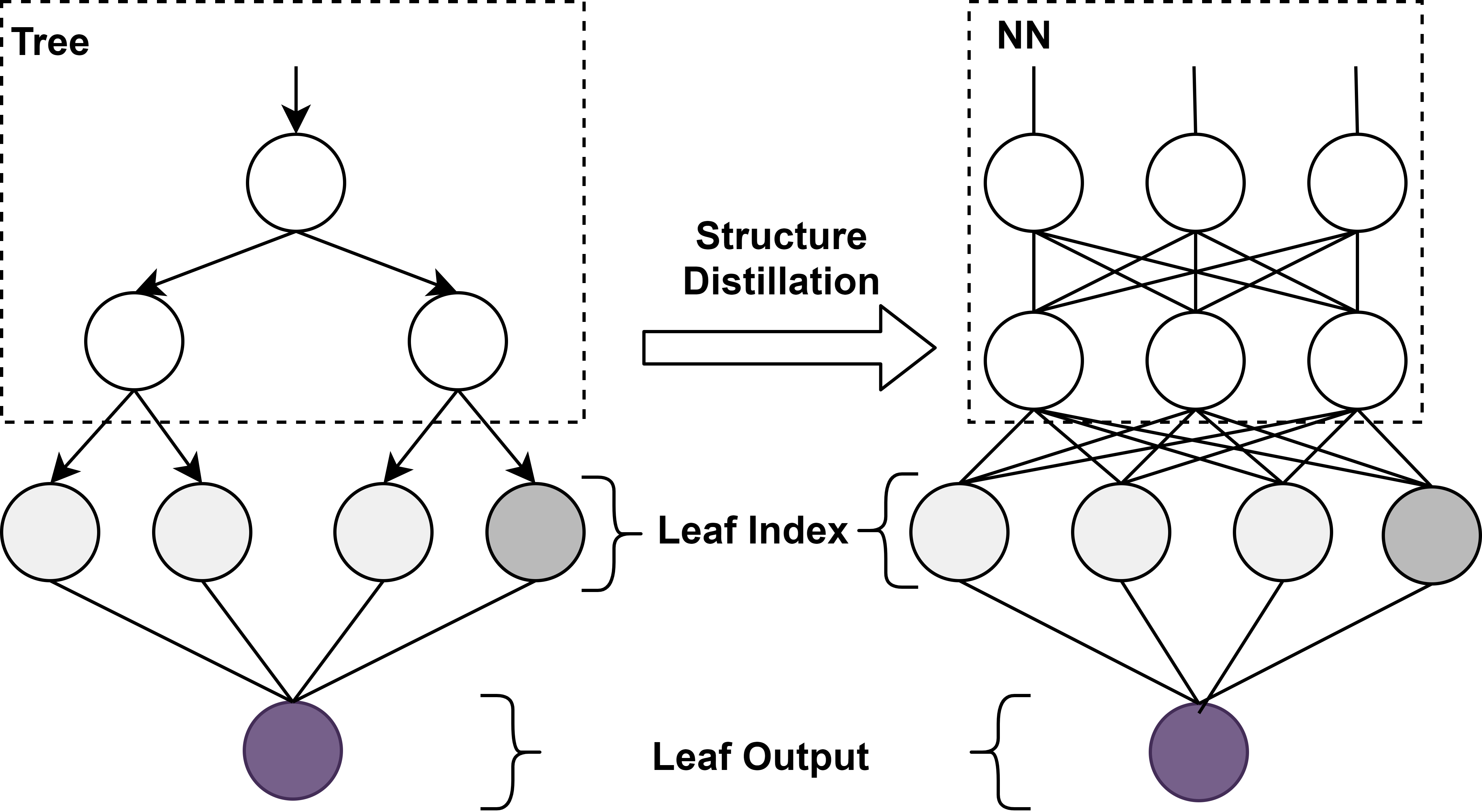}
\caption{Tree structure distillation by leaf index.}
\label{fig:GBDT2NN1}
\end{figure}

GBDT2NN converts GBDT into the NN model by learning the tree structure. As illustrated in Fig.\ref{fig:GBDT2NN1}, GBDT2NN approximates the structure function of decision tree by fitting the cluster results produced by the tree, and the leaf index is taken as the cluster results. For a single tree $t$, it denotes the one-hot representation of leaf index for sample $\mathbf{x}$ as $L_{t}(\mathbf{x})$, and the NN part learns the mapping function between $\mathbf{x}$ and $L_{t}(\mathbf{x})$ as:
\begin{equation}
     \min_{\theta_\text{pred}} \mathbb{E}_{\mathbf{x}} \big[ \mathit{l}_1 ( \mathit{NN} (\mathbf{x}; \theta_\text{pred}), L_t(\mathbf{x}) ) \big],
\label{eq:3}
\end{equation}
where $\mathit{NN}(\cdot; \theta_\text{pred})$ is a neural network parameterized by $\theta_\text{pred}$, and can be optimized by back-propagation, $\mathit{l}_1$ is a loss function for the prediction task. And the output of NN corresponding to tree $t$ can be denoted as:
\begin{equation}
\hat{y}_t(\mathbf{x}) = \mathit{NN}(\mathbf{x}; \theta_\text{pred}) \cdot q_t
\end{equation}
where $q^t$ is the leaf values of tree $t$.

\begin{figure}[htbp]
\centering
\includegraphics[width=0.8\textwidth]{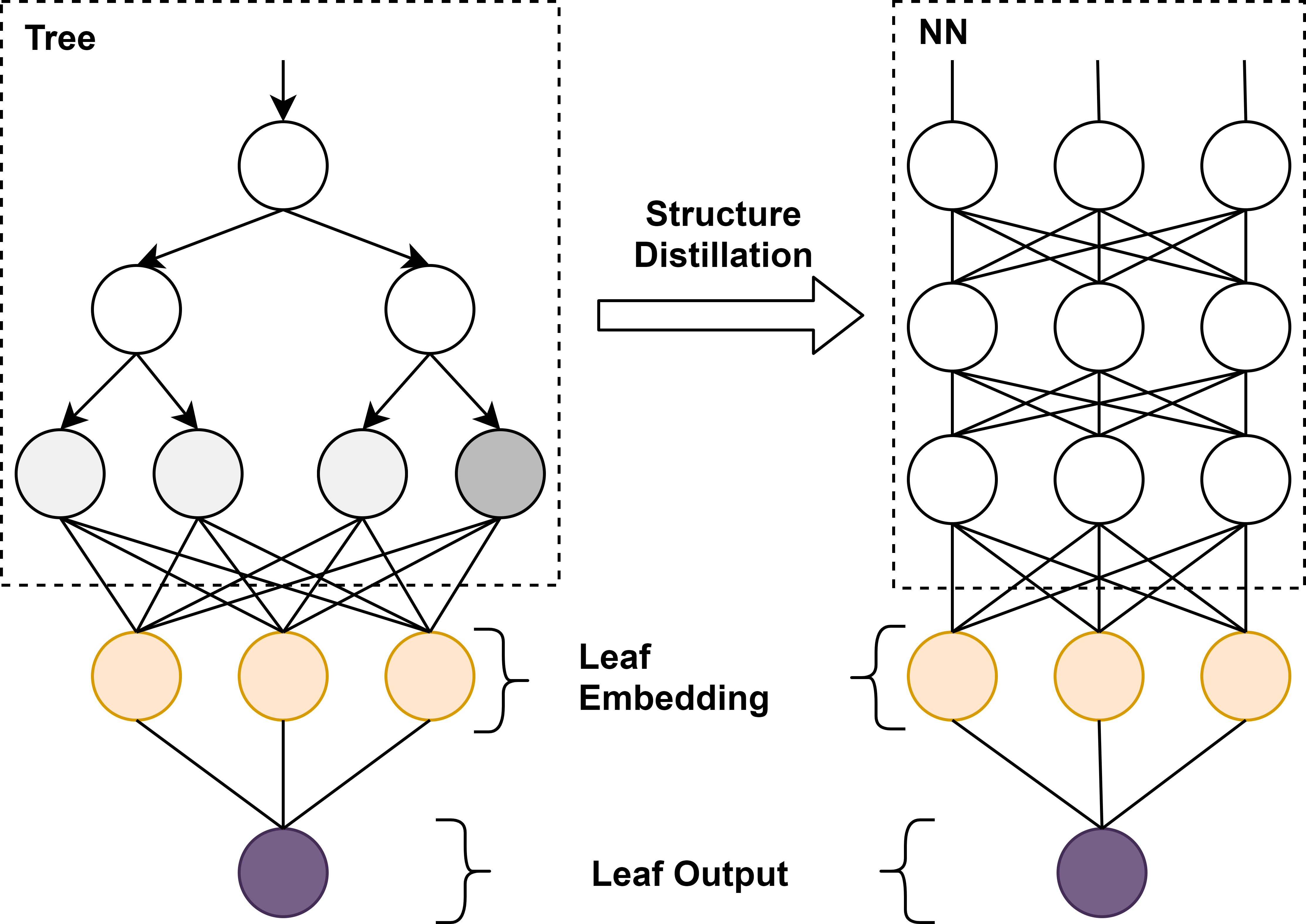}
\caption{Tree Structure distillation by leaf embedding.}
\label{fig:GBDT2NN2}
\end{figure}

To improve the efficiency, GBDT2NN adopts {\em Tree-Selected-Features}, {\em Leaf Embedding Distillation} and {\em Tree Grouping} strategies to reduce the input features, high dimensional one-hot representation $L$ and $\#\mathit{NN} = \#\mathit{tree}$ NN models.
\begin{enumerate}
    \item {\em Tree-Selected-Features}: Rather than using all input features, NN only uses tree-selected features as the inputs. The indices of the selected features are defined as $\mathbb{I}_t$ in the tree $t$, and the input can be denoted as $\mathbf{x}(\mathbb{I}_t)$.
    \item {\em Leaf Embedding Distillation}: As shown in Fig.\ref{fig:GBDT2NN2}, it adds an embedding layer between leaf indices and output. The learning process of embedding can be written as:
        \begin{equation}
        \min_{w,  w_0, w_t} \mathbb{E}_\mathbf{x}  [ \mathit{l}_1 (w^{\top}  \mathcal{H}( L_t(\mathbf{x}); w_t) + w, p_t(\mathbf{x}))],
        \end{equation}
        where $H_t(\mathbf{x}) = \mathcal{H}(L_t(\mathbf{x}); w_t)$ is a special type of fully connected network which converts $L_t(\mathbf{x})$ to the dense embedding, $p_t(\mathbf{x})$ is the predict leaf value of sample $\mathbf{x}$. After that, instead of one-hot representation $L$, GBDT2NN takes the dense embedding $\mathcal{H}$ as the target to approximate the function of tree structure to get $\mathit{NN}(\mathbf{x}(\mathbb{I}_t); \theta_\text{pred})$ as in Equation \ref{eq:3}.
    \item {\em Tree Grouping}: GBDT2NN use the equally randomly grouping, and extends the {\em Leaf Embedding Distillation} for each group of trees. For the tree group $\mathbb{T}$, the leaf embedding process is as:
        \begin{equation}
        \min_{w, w_0, w_{\mathbb{T}}} \mathbb{E}_\mathbf{x} [\mathit{l}_1 (w^{\top} \mathcal{H}( \|_{t\in \mathbb{T}} (L_t(\mathbf{x})); w_{\mathbb{T}}) + w_0, \sum_{t \in \mathbb{T}} p_t(\mathbf{x}))]
        \label{eq:4}
        \end{equation}
        where $\|(\cdot)$ is the concatenate operation, and $H_{\mathbb{T}}(\mathbf{x}) =\mathcal{H}(\|_{t\in \mathbb{T}} (L_t(\mathbf{x})); w_{\mathbb{T}})$ converts the multiple one-hot leaf index vectors to the dense embedding. After that, GBDT2NN takes the new dense embedding as the target to approximate the function of tree structure as in Equation \ref{eq:3}.
\end{enumerate}

Finally, the output of the GBDT2NN is the the summation of all tree groups.

\section{Proposed Explaining Methods for GBDT2NN}
\label{sec:distill}


In this section, we bring the path concept into the interpretation model, and propose a new idea to give explanations of the individualized sample for GBDT2NN.
Based on the new idea, we put forward two concrete methods: independent method and joint method.

\subsection{Intuition for the explanations of GBDT2NN}

Each leaf index $s$ in the teacher's model represents a decision path $\pi_s$, as well as a prediction $y_s$. As discussed above, GBDT2NN learns the leaf indices of samples before training the NNs to retain the structure functions of decision tree, so each embedding representation $L$ represents a prediction $y_L$. Any given sample $\mathbf{x}$ would go through the internal nodes or neurons, and arrives at one leaf index $s(x)$ or generates an embedding vector $L(x)$, and the intermediate states determine the output directly.

On the other hand, the structured-based interpreting methods of GBDT explain the single sample's prediction by analyzing the features' gain of the passing path.
And we can get the interpretation vector $\Phi(\mathbf{x})$ of sample $\mathbf{x}$, which actually reflects the attributes of features used in decision path $\pi_{s(x)}$ that sample $\mathbf{x}$ goes through. It is also means that the feature attributes of every sample can also be determined by the leaf node. According to the effects of the embedding layer and the additive relationship of NNs in GBDT2NN, we take use of the different embedding of samples to give the explanations. And the attributes reflected by the embedding of GBDT2NN can be leant from the explanations of teachers' model. Therefore, rather than learning the relations between input vector of $\mathbf{x}$ and interpretation vector $\hat{\Phi}(\mathbf{x})$, we introduce the embedding concept and divide the original problem into a two-stage framework:
\begin{enumerate}
    \item The mapping between input vector of $\mathbf{x}$ and embedding representation $L(x)$;
    \item The relations between embedding representation $L(x)$ and the final interpretation vector $\hat{\Phi}(\mathbf{x})$.
\end{enumerate}

\subsection{The independent method}
In this section, we put forward an independent interpretable model for GBDT2NN.
The introduction also starts with the learning of a single NN $t$ of GBDT2NN. To retain the ability of online update, we use NN to approximate the first mapping function between $\mathbf{x}$ and $L_{t}(\mathbf{x})$ as in Equation \ref{eq:3}.
After that, we can get the NN model $\mathit{NN}(\cdot; \theta_\text{pred})$. For the second learning process, we use the interpretation vectors $\Phi(\mathbf{x})$ of GBDT to learn the mapping from the leaf index to the final interpretation $\hat{\Phi}(\mathbf{x})$, whose objective can be described by Equation \ref{eq:post-hod-l2}:
\begin{equation}
     \min_{w, w_0}\mathbb{E}_{\mathbf{x}} \big[ \mathit{l}_2 ( w^\top \mathit{NN} (\mathbf{x(\mathbb{I}_t)}; \theta_\text{pred}) + w_0, \Phi_t(\mathbf{x}) ) \big],
    \label{eq:post-hod-l2}
\end{equation}
where $w$ is a transition matrix that converts the one-hot leaf index to the interpretation vector, $w_0$ is the bias vector, $l_2$ is the loss function of the interpretation task, which is translated into a regression problems with MSE(mean-square error) to fit dense interpretation vectors $\Phi_t(\mathbf{x})$.
After that, we can get the interpretation vector $\hat{\Phi}_t (\mathbf{x})$ of the NN $t$ for sample $\mathbf{x}$, as follows
\begin{equation}
    \hat{\Phi}_t (\mathbf{x}) = w^\top \mathit{NN} (\mathbf{x(\mathbb{I}_t)}; \theta_\text{pred}) + w_0 .
\end{equation}

The interpretation of ensemble NNs is obtained by summing up of every single NN attribution vector:
\begin{equation}
    \Phi_\text{ensemble} (\mathbf{x})= \sum_{t=1}^T \hat{\Phi}_t (\mathbf{x}),
\end{equation}
where $T$ is the total number of NNs.

\begin{figure}[htbp]
\centering
\includegraphics[width=0.8\textwidth]{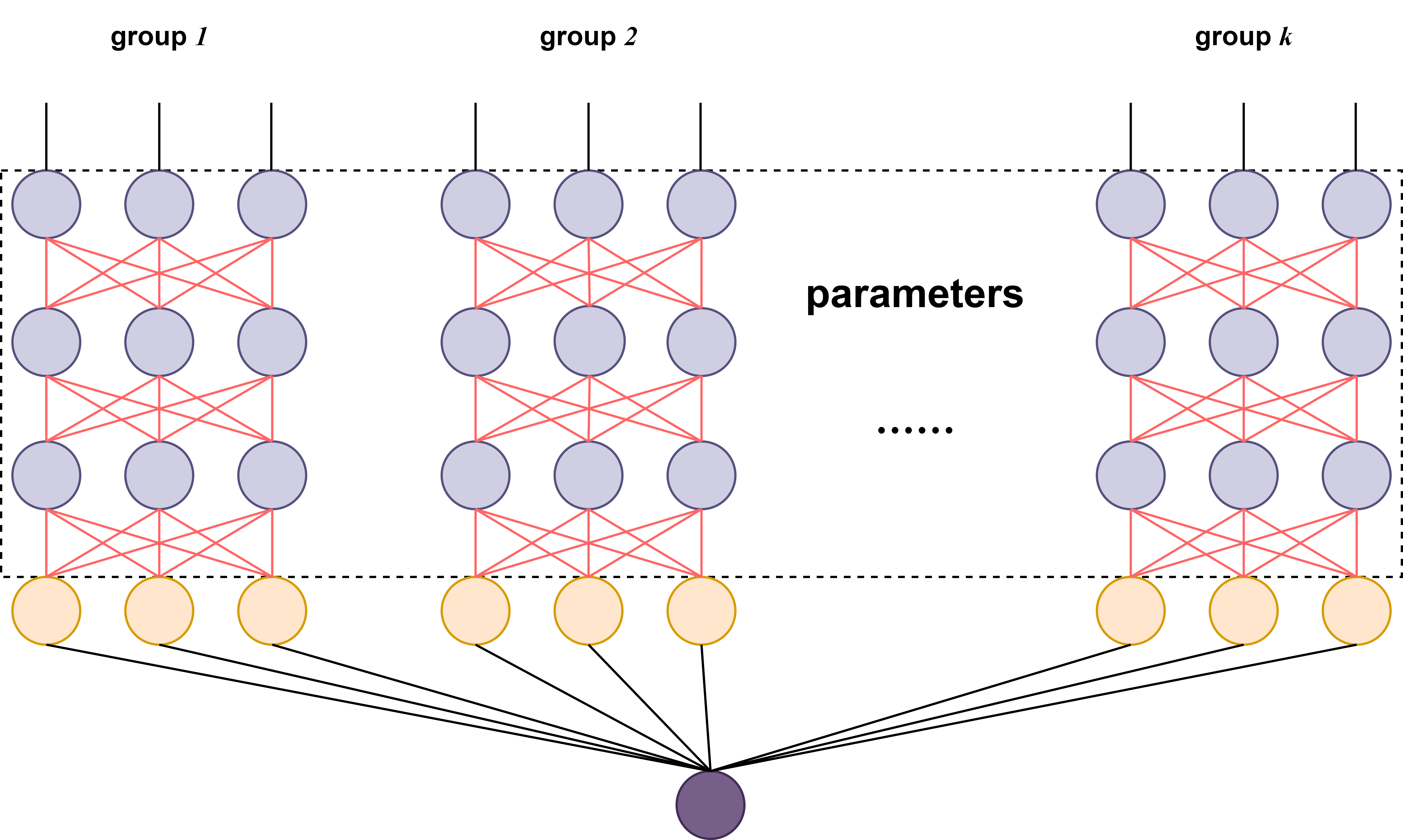}
\caption{The parameters of independent method loaded from GBDT2NN.}
\label{fig:gbdt2nn}
\end{figure}

Since the above procedure is the same as GBDT2NN, as well as the additive relations of the NNs' interpretation vectors, the efficiency strategies for GBDT2NN, such as {\em Tree-Selected-Features}, {\em Leaf Embedding Distillation} and {\em Tree Grouping}, are also suitable for the out proposed method. Therefore, the interpretable model does not need to train the first process again, only needs to load the parameters from input to the embedding layer of GBDT2NN, which is already trained for prediction. Parameters that are loaded from the GBDT2NN model are shown in Fig. \ref{fig:gbdt2nn}.

Then we get $k$ NNs which are corresponding to $k$ groups of trees in GBDT2NN. We define $\mathbf{x}[\mathbb{T}]$ to be the features' subset that are used by the tree group $\mathbb{T}$, and denote
\begin{equation}
G_\mathbf{x} = \|_{j \in k} \mathit{NN} (\mathbf{x}[{\mathbb{T}_j}]  ; \theta^{\mathbb{T}_j}_\text{pred} )
\end{equation}
as the concatenate of multi-group embedding for all $k$ NNs.
Then, we can fit the affine function mapping from the embedding to the interpretation vector, such that
\begin{equation}
    \min_\mathbf{\theta} \mathbb{E}_{\mathbf{x}} \big[ l_2\big( w^\top G_\mathbf{x} + w_0, \Phi(\mathbf{x})) \big],
\label{eq:2}
\end{equation}
and the learnt interpretation vector by the new method is
\begin{equation}
\Phi_\text{ensemble}(\mathbf{x}) =  w^\top G_\mathbf{x} + w_0 .
\end{equation}

\begin{figure}[htbp]
\centering
\includegraphics[width=0.8\textwidth]{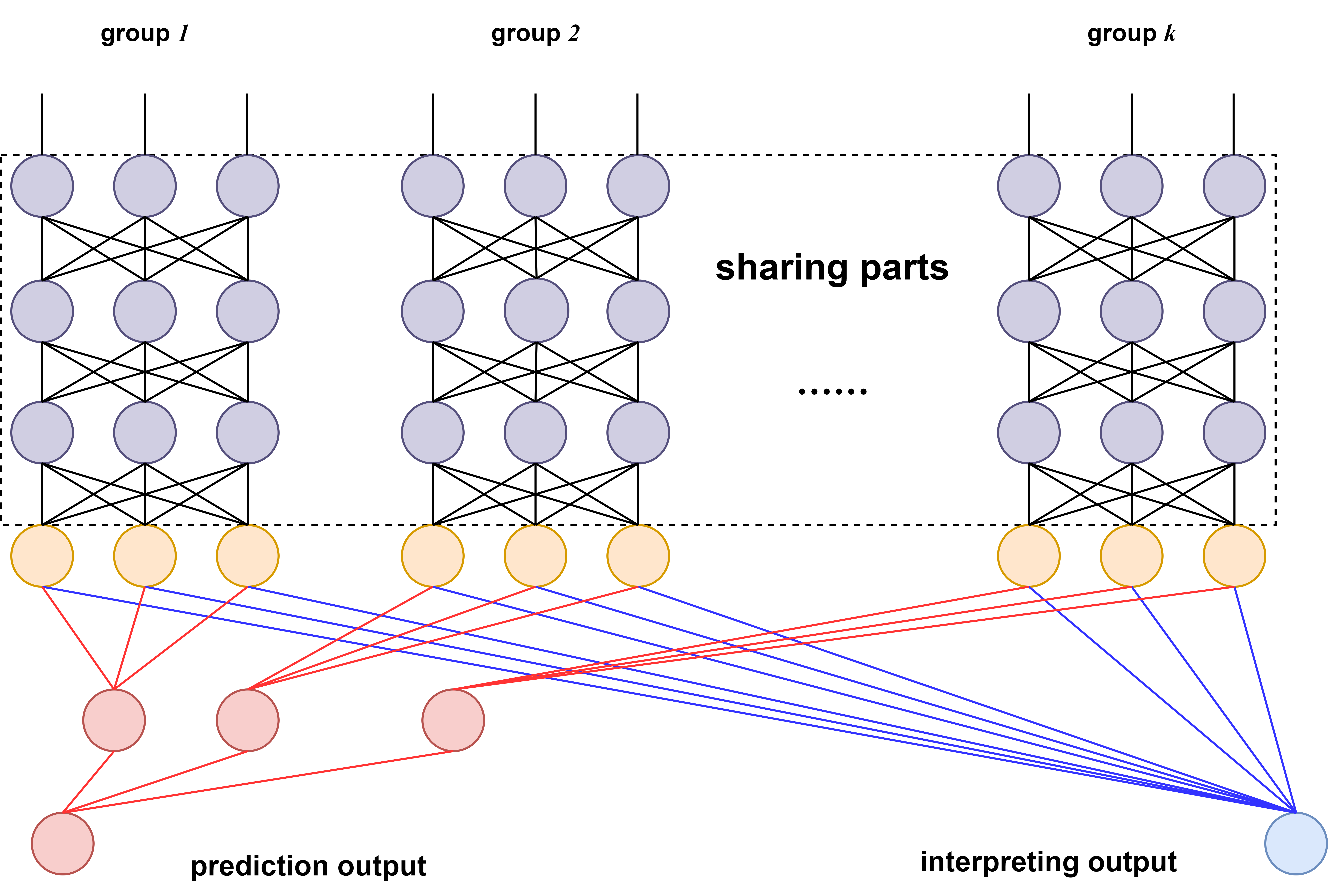}
\caption{The mixed model with two outputs.}
\label{fig:two_ouput}
\end{figure}

Then we obtain explanation model, and the prediction model shares the same parameters with the explanation model except the final fully connected layer. Furthermore, we merge two models and offer a mixed framework with two outputs, as shown in Fig. \ref{fig:two_ouput}.

From \cite{ke2019deepgbm}, the online update operations of GBDT2NN only adjust the parameters from input to the embedding layer, which are actually the parameter-sharing parts of the two models as shown in Fig.\ref{fig:two_ouput}. That also means the online update operations only "recombine" the embedding, and retain the mapping from embedding layers to the final output. Therefore, our proposed method is also applicable to explain the updated GBDT2NN model.

\subsection{The joint method}

In the two-step learning process of GBDT2NN model, the embedding layer is learned from the leaf index and then used as the distillation target of NN model.
Better representation of the embedding leads to better approximation of the tree structure. Inspired by this, we attempt to incorporate the interpretation process into the construction of the prediction model, and train the predictive model and interpreting model jointly.
Compared with leaf indices, structure-based interpretations directly reflects the feature importance of the decision paths, and taking this detailed information as another target can improve the embedding learning process.
Therefore, we propose a two-view loss function to combine leaf indices and interpreting vectors.
The learning process of embedding can denote as
\begin{equation}
\begin{aligned}
    \min_{w_1,  w_2, w_{0_1}, w_{0_2}, w_{\mathbb{T}_j}} & \mathbb{E}_\mathbf{x}  [ \lambda \cdot \mathbb{E}_{j} [ \mathit{l}_1 (w_1^{\top}  \mathcal{H}( \|_{t\in \mathbb{T}_j} (L_t(\mathbf{x})); w_{\mathbb{T}_j}) + w_{0_1}, \sum_{t \in \mathbb{T}_j} p_t(\mathbf{x}))] \\
    + &(1 - \lambda) \cdot \mathit{l}_2 (w_2^{\top}  \|_{j \in k} (\mathcal{H}( \|_{t\in \mathbb{T}_j} (L_t(\mathbf{x})); w_{\mathbb{T}_j})) + w_{0_2}, \Phi(\mathbf{x}) )],
\end{aligned}
\end{equation}
where $\lambda$ is a parameter that controls the trade-off between two parts. The first part measures the prediction accuracy as in Equation \ref{eq:4}, and the latter part evaluates the interpreting quality as in Equation \ref{eq:2}. After getting the embedding, the follow-up process is the same as that of GBDT2NN model and the above proposed method.

\section{Experiments}
\label{sec:exp}

To evaluate the performance of the proposed methods, we conduct experimental comparisons on several public datasets.
Specifically, OnlineNewsPop\footnote{https://archive.ics.uci.edu/ml/machine-learning-databases/Online+News+Popularity} is a standard dataset obtained from the UCI machine learning repository, and we convert it to a classification problem by setting the popular value of "shares" to 1400 as in \cite{fernandes2015proactive}.
AutoML-3\footnote{https://www.4paradigm.com/competition/nips2018} is from "AutoML for Lifelong Machine Learning" Challenge in NeurIPS 2018.
MNIST\footnote{http://yann.lecun.com/exdb/mnist/} is a database for handwritten digit recognition task. We convert the original dataset to several binary classification datasets. More specifically, we choose two binary classificaiton tasks, number 0 and number 4, as the former is easy to learn and the latter is much more difficult.
Zillow\footnote{https://www.kaggle.com/c/zillow-prize-1} is a dataset from Kaggle competition.
The experimental datasets contain classification and regression problems, and the quantities are varying from large to small. The train set and test set are constructed according to the existing literatures.
The statistics of the datasets are shown in Table \ref{tbl:mnist-exp}.

\begin{table}[!t]
\caption{Statistics of the experimental datasets}
\label{tbl:mnist-exp}
\centering
\begin{tabular}{c|ccccc}
\hline
Dataset & 	Sample Size & Train Size & Test Size & 	No. Features 	& 	Task \\
\hline
AutoML-3 	& 	0.78M & 0.70M & 0.08M & 	71 	& 	Cls \\
\hline
OnlineNewsPop   &  39.6K & 30K & 9.6K &   58  &   Cls\\
\hline
MNIST-0 	& 	70K & 60K & 10K & 	28 * 28 	& 	Cls\\
\hline
MNIST-4 	& 	70K & 60K & 10K & 	28 * 28 	&  	Cls\\
\hline
Zillow 	& 	90.3K & 81.3K & 9K & 	57 	& 	Reg\\
\hline
\end{tabular}
\end{table}


\subsection{Evaluation Metrics}

Since consensus on the measurement of interpreting quality though  has not yet arrived, this paper takes the results obtained from the original model GBDT as our "baseline", and use NDCG \cite{jarvelin2002cumulated} and top-k coverage to measure the correlation between the two explanation results. NDCG is a measure of ranking quality, and it is used to evaluate the ranking quality of our proposed method compared with the "baseline".
$c_k(\mathbf{x})$ is a variant of top-k hit rate, can be mathematically expressed as:
\begin{equation}
c_k(\mathbf{x}) = \frac{\sum_{j \in F} \mathbb{I} (j \in \Phi_\text{top-k}(\mathbf{x})) \cdot \mathbb{I} (j \in \hat{\Phi}_\text{top-k}(\mathbf{x}))} {k},
\end{equation}
where $F$ is the union of used features by all NNs, $\Phi_\text{top-k}(\mathbf{x})$ is the top $k$ most important features of GBDT obtained by structure-based tree explanation method, $\hat{\Phi}_\text{top-k}(\mathbf{x})$ is the top $k$ important features of GBDT2NN obtained by our proposed methods.
Top-k coverage $c_k$ is aimed to calculate the intersect rate of the important features.

Besides above quantitative evaluations, we also adopt the intuitive representation to evaluate experiment results, which is only applicable for the image datasets. As for GBDT and GBDT2NN, the pixel points are actually the input features of models. The specific implement is to show the pixel points which contribute much to classification, and we call it visual evaluation.

To evaluate the prediction improvements of the joint method, we use AUC to measure the prediction results of classification datasets, and use MSE to measure that of regression datasets.

\subsection{Structured-based Interpretation Method of GBDT}
\label{sec:method_setting}
\begin{figure}
\centering
\subfigure[number 0]{
    \label{fig:subfig:p} 
    \includegraphics[width=1.5in]{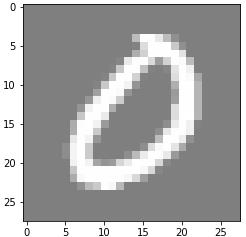}}
\subfigure[number 4]{
    \label{fig:subfig:p} 
    \includegraphics[width=1.5in]{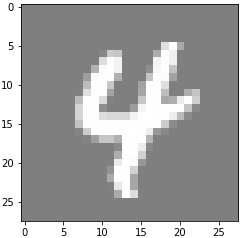}}
  \caption{The original figures of digit 0 and 4.}

  \label{fig:original figures} 
\end{figure}

\begin{figure*}
  \centering
  \subfigure[Top-20 important features.]{
    \includegraphics[width=1.5in]{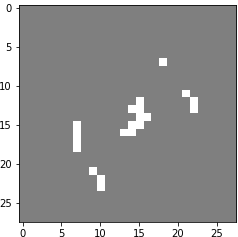}}
  \hspace{0.1in}
  \subfigure[Top-50 important features.]{
    \includegraphics[width=1.5in]{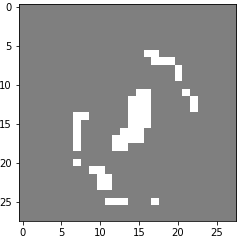}}
  \hspace{0.1in}
  \subfigure[Top-100 important features.]{
    \includegraphics[width=1.5in]{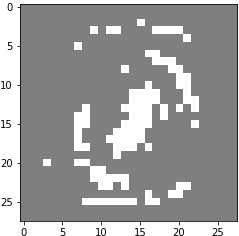}}
  \\
  \subfigure[Top-20 important features in outline.]{
    \includegraphics[width=1.5in]{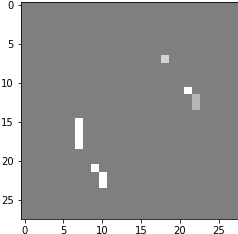}}
  \hspace{0.1in}
  \subfigure[Top-50 important features in outline.]{
    \includegraphics[width=1.5in]{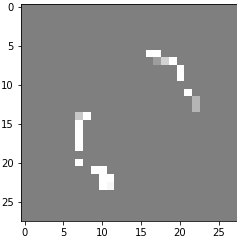}}
  \hspace{0.1in}
  \subfigure[Top-100 important features in outline.]{
    \includegraphics[width=1.5in]{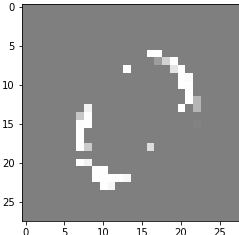}}
  \caption{The evaluation result of digit 0 with Sabbas. The first row are the top-k important features, and we use the white pixels to represent them. And the second row shows the top-k important features in the digits' outline. The values of k are set as 20, 50 and 100.}

  \label{fig:Sabbas_0} 
\end{figure*}

\begin{figure*}
  \centering
  \subfigure[Top-50 important features.]{
    \includegraphics[width=1.5in]{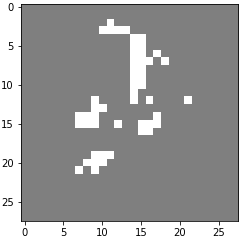}}
  \hspace{0.1in}
  \subfigure[Top-100 important features.]{
    \includegraphics[width=1.5in]{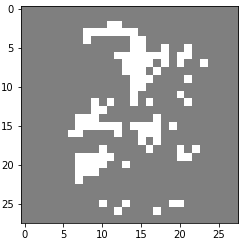}}
  \hspace{0.1in}
  \subfigure[Top-150 important features.]{
    \includegraphics[width=1.5in]{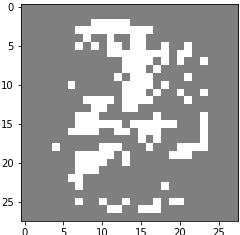}}
  \\
  \subfigure[Top-50 important features in outline.]{
    \includegraphics[width=1.5in]{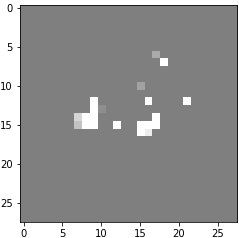}}
  \hspace{0.1in}
  \subfigure[Top-100 important features in outline.]{
    \includegraphics[width=1.5in]{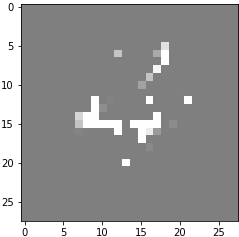}}
  \hspace{0.1in}
  \subfigure[Top-150 important features in outline.]{
    \includegraphics[width=1.5in]{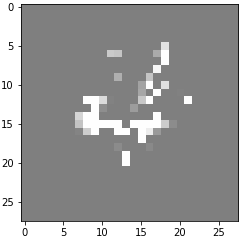}}
  \caption{The evaluation result of digit 4 with Sabbas. The first row are the top-k important features, and we use the white pixels to represent them. And the second row shows the top-k important features in the digits' outline. The values of k are set as 50, 100 and 150.}

  \label{fig:Sabbas_4} 
\end{figure*}

As mentioned in \ref{sec:distill}, our new method builds the explanation model based on the structured-based interpretation method of GBDT. Here, we use Sabbas to get the features importance of each path. To illustrate the effective of Sabbas, we represent the results of MNIST dataset using visual evaluation method, and the original figures are shown in Fig. \ref{fig:original figures}. As in Fig. \ref{fig:Sabbas_0} and Fig. \ref{fig:Sabbas_4}, each figure contains two classes results. The first row are the top-k important features learnt by Sabbas, where we use the white pixels to represent them. And the second row replace the color of the important features with the color of the original pictures. To make the results more obvious, we set the background gray values the same with that of the original pictures, and the figure actually shows the top-k important features in the digits' outline. From these, it is easy to see that Sabbas can distinguish the region of the handwritten digits from the background, and grasp the outline pixels. And the result of digit 0 is a little better than 4, corresponding that the classification accuracy of digit 0 is better than that of 4. The results of visual evaluation prove that the Sabbas is effective, considering that GBDT is not so good at dealing with image problems.

\subsection{Experiment Results}

Here, we present the performance of the independent interpretable method. The results of NDCG and AVG($c_k$) are shown in Table \ref{tbl:eval-ndcg} and Table \ref{tbl:eval-topk}. From them we can draw the following observations. First, the repeatability of the two explanations is high, although the important features obtained by our method from GBDT2NN are some different from the top attributes of GBDT. Compared with AVG($c_k$), the values of NDCG are higher, which means that the relative rankings of top attributes derived from the two models are relatively consistent. Besides, the evaluation values of digit 4 is ``worse'' than that of digit 0.

\begin{table}[!t]
\centering
\caption{The results of NDCG. NDCG is a measure of ranking quality, and it is used to evaluate the ranking quality of our proposed method compared with that of GBDT}
\label{tbl:eval-ndcg}
\begin{tabular}{l|cc}
\hline
Dataset & NDCG@5 & NDCG@10 \\
\hline
AutoML-3 & 0.9466 & 0.9305 \\
\hline
OnlineNewsPop & 0.8396 & 0.8316 \\
\hline
MNIST-0 & 0.8775 & 0.8629 \\
\hline
MNIST-4 & 0.7506 & 0.7534 \\
\hline
Zillow & 0.7614 & 0.7564 \\
\hline
\end{tabular}
\end{table}

\begin{table}[!t]
\centering
\caption{The results of AVG($c_k$). $c_k$ is aimed to calculate the intersect rate of the important features between different interpretable methods.}
\label{tbl:eval-topk}
\begin{tabular}{l|cccc}
\hline
Dataset & AVG($c_1$) & AVG($c_3$)& AVG($c_5$) & AVG($c_{10}$) \\
\hline
AutoML-3 & 0.7861 & 0.7895 & 0.8121 & 0.7197 \\
\hline
OnlineNewsPop & 0.6682 & 0.5883 & 0.5601 & 0.5895 \\
\hline
MNIST-0 & 0.7998 & 0.7904 & 0.65934 & 0.52588 \\
\hline
MNIST-4 & 0.4732 & 0.5596 & 0.53944 & 0.54203 \\
\hline
Zillow & 0.5063 & 0.5468 & 0.5667 & 0.5836 \\
\hline
\end{tabular}
\end{table}

\begin{figure*}
  \centering
  \subfigure[Top-20 important features.]{
    \includegraphics[width=1.5in]{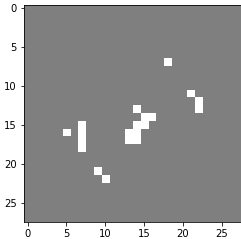}}
  \hspace{0.1in}
  \subfigure[Top-50 important features.]{
    \includegraphics[width=1.5in]{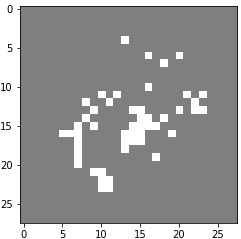}}
  \hspace{0.1in}
  \subfigure[Top-100 important features.]{
    \includegraphics[width=1.5in]{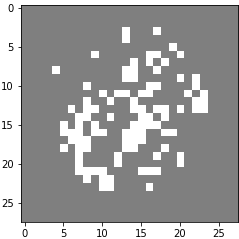}}
  \\
  \subfigure[Top-20 important features in outline.]{
    \includegraphics[width=1.5in]{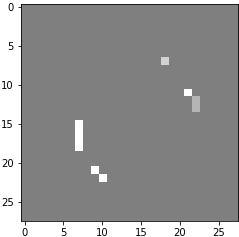}}
  \hspace{0.1in}
  \subfigure[Top-50 important features in outline.]{
    \includegraphics[width=1.5in]{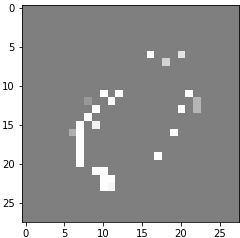}}
  \hspace{0.1in}
  \subfigure[Top-100 important features in outline.]{
    \includegraphics[width=1.5in]{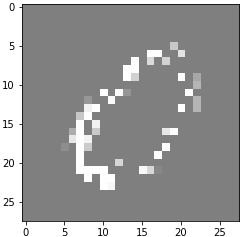}}
  \caption{The evaluation result of digit 0 with independent method. The first row are the top-k important features, and we use the white pixels to represent them. And the second row shows the top-k important features in the digits' outline. The values of k are set as 20, 50 and 100.}

  \label{fig:PostHoc_0} 
\end{figure*}

\begin{figure*}
  \centering
  \subfigure[Top-50 important features.]{
    \includegraphics[width=1.5in]{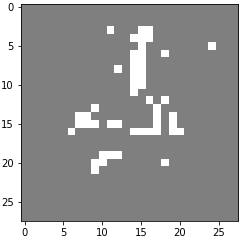}}
  \hspace{0.1in}
  \subfigure[Top-100 important features.]{
    \includegraphics[width=1.5in]{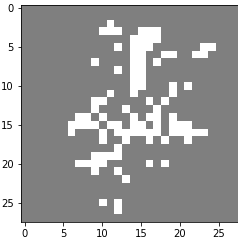}}
  \hspace{0.1in}
  \subfigure[Top-150 important features.]{
    \includegraphics[width=1.5in]{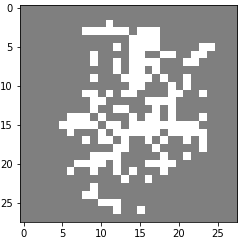}}
  \\
  \subfigure[Top-50 important features in outline.]{
    \includegraphics[width=1.5in]{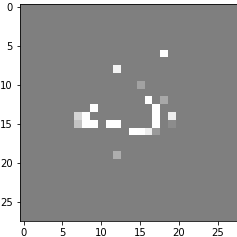}}
  \hspace{0.1in}
  \subfigure[Top-100 important features in outline.]{
    \includegraphics[width=1.5in]{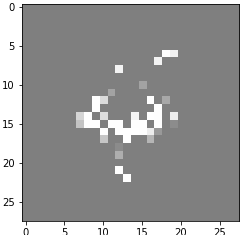}}
  \hspace{0.1in}
  \subfigure[Top-150 important features in outline.]{
    \includegraphics[width=1.5in]{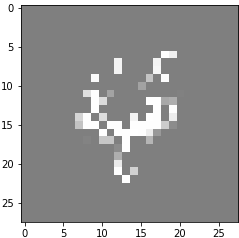}}
  \caption{The evaluation result of digit 4 with the independent method. The first row are the top-k important features, and we use the white pixels to represent them. And the second row shows the top-k important features in the digits' outline. The values of k are set as 50, 100 and 150.}

  \label{fig:PostHoc_4} 
\end{figure*}

The quantitative evaluations only measure the correlations between two methods. To better know the explanation results of the proposed method, we also plot the visual evaluation results of MNIST-0 and MNIST-4 in Fig. \ref{fig:PostHoc_0} and Fig. \ref{fig:PostHoc_4}. Compared with the results in \ref{sec:method_setting}, the figures in Fig. \ref{fig:PostHoc_0} and Fig. \ref{fig:PostHoc_4} present a more accurate feature description of the digits. Especially for the digit 0, our method not only grasps the outline better, and it learns the background part inside the circle of digit 0, which is also important for the recognition. Note that, the white pixels in the latter three pictures of Fig. \ref{fig:PostHoc_0} and Fig. \ref{fig:PostHoc_4} seem a little more than in Fig \ref{fig:Sabbas_0} and Fig \ref{fig:Sabbas_4}, and that is because that Sabbas picks more background pixels as important features where the background gray is the same with the original picture. Therefore, these results prove the effectiveness of our proposed idea and method.

\begin{figure*}[htbp]
\centering
\begin{minipage}{0.3\linewidth}
  \centerline{\includegraphics[width=\textwidth]{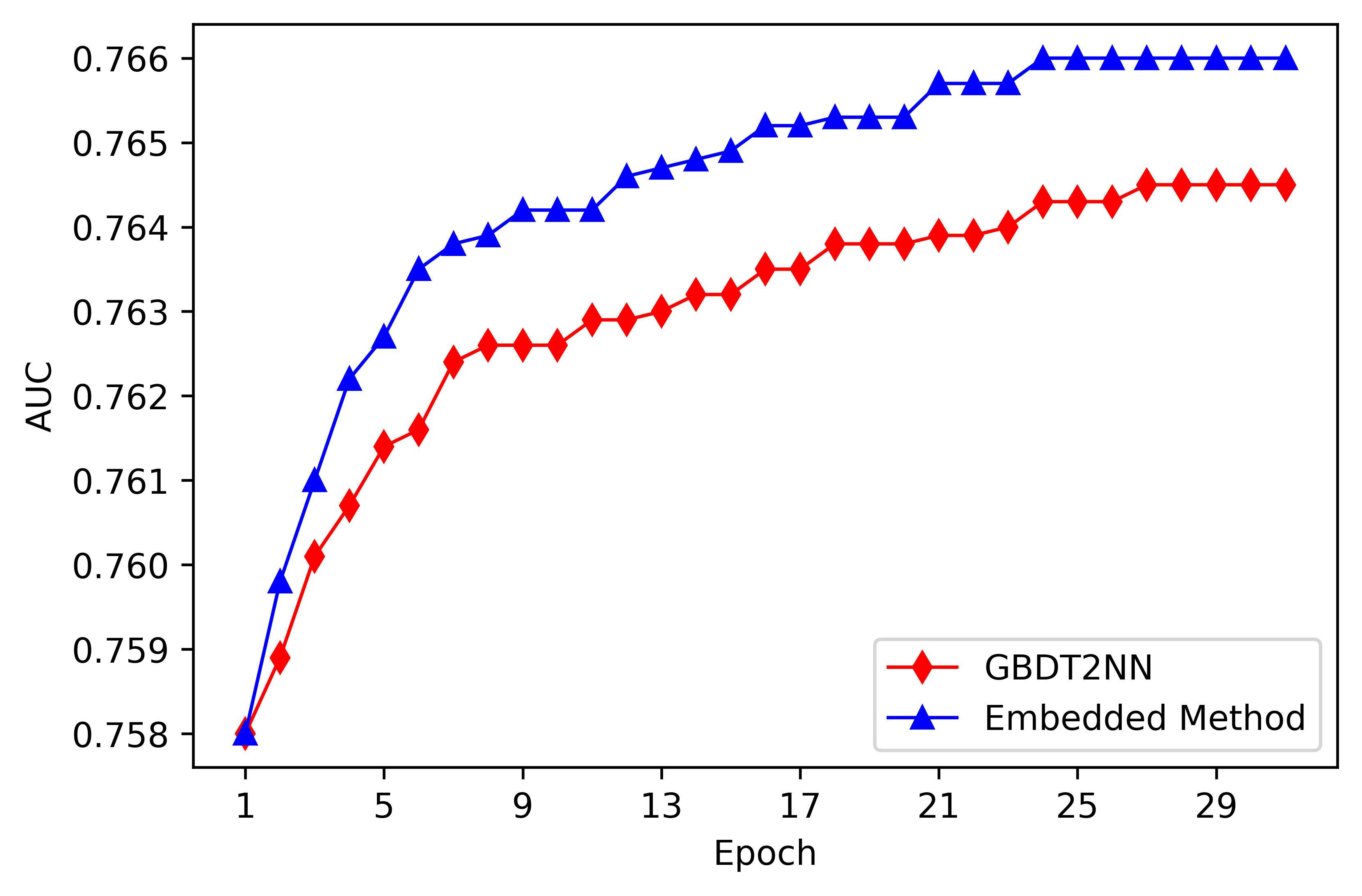}}
  \centerline{AutoML-3.}
\end{minipage}
\hfill
\begin{minipage}{0.3\linewidth}
  \centerline{\includegraphics[width=\textwidth]{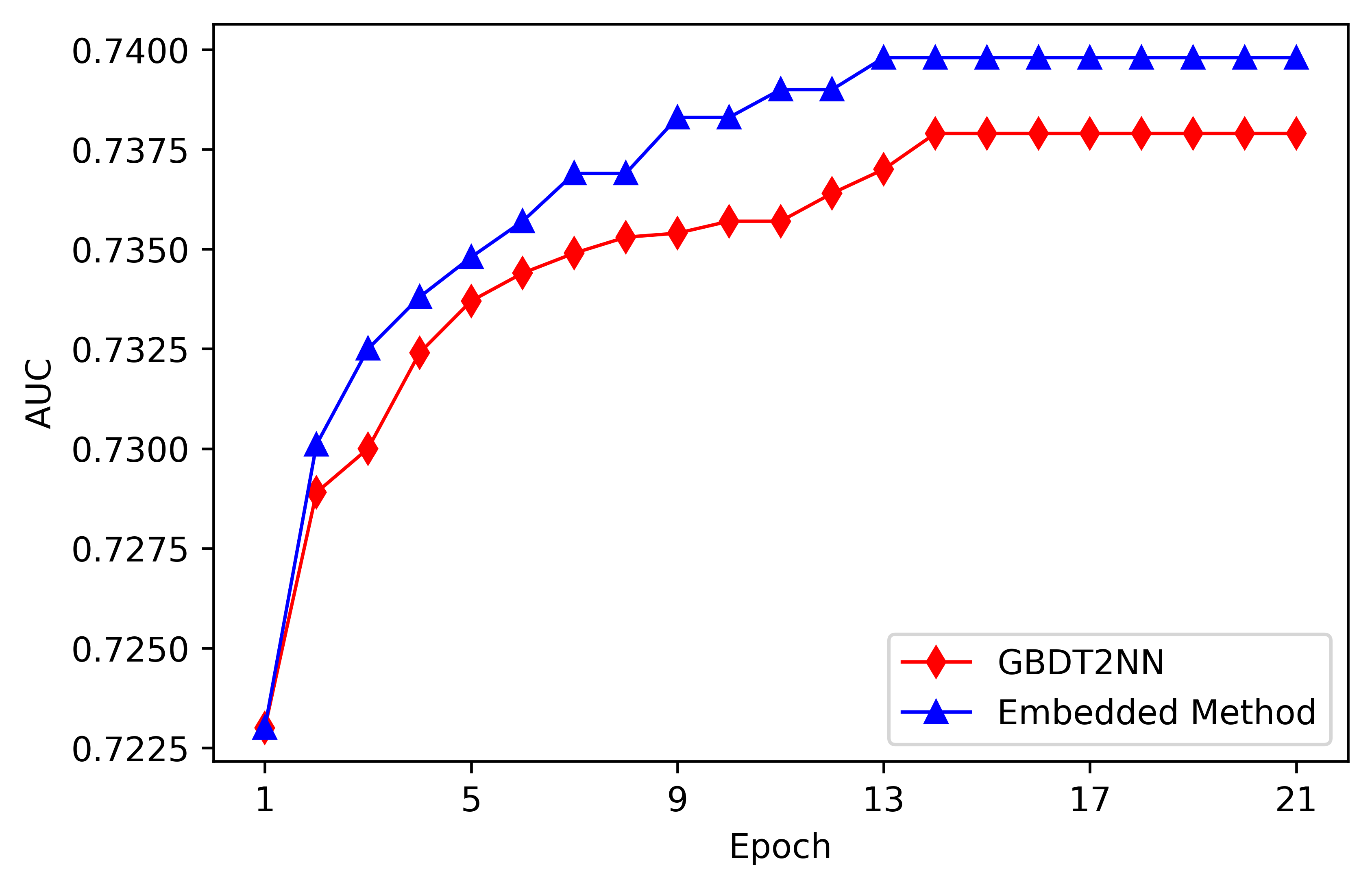}}
  \centerline{OnlineNewsPop.}
\end{minipage}
\hfill
\begin{minipage}{0.3\linewidth}
  \centerline{\includegraphics[width=\textwidth]{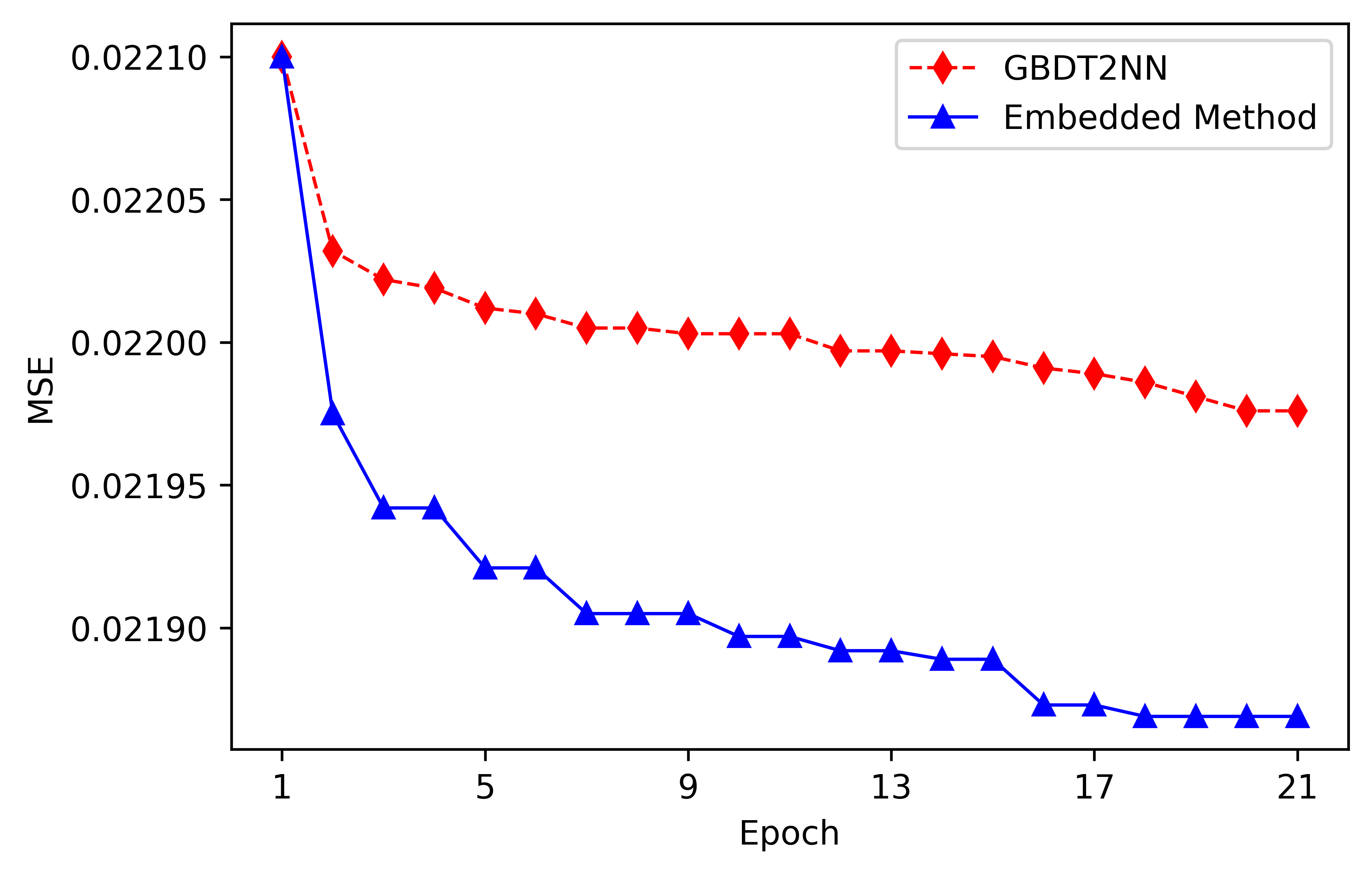}}
  \centerline{Zillow.}
\end{minipage}

\caption{Epoch-Loss curves over test data. The loss of AutoML-3 and OnlineNewsPop is AUC, and the loss of Zillow is MSE.}
\label{fig:loss-epoch}
\end{figure*}

Besides the independent interpretable method, we also propose an joint method. In theory, the joint method can improve the prediction performance while providing the interpretation results. We only conduct experiments on several datasets, and the prediction results are shown in Table \ref{tbl:pred}. The interpreting results are almost the same as the independent method, which are omitted. From Table \ref{tbl:pred}, we can see that the prediction performance of GBDT2NN with our proposed two-view loss function has improved. To investigate the impacts of the joint method on the convergence speed, we represent the performance of loss on the test data with increasing epochs in Fig. \ref{fig:loss-epoch}. Specifically, the loss function of AutoML-3 and OnlineNewsPop is AUC, and the loss function of Zillow is MSE. From these figures, we can find that the model trained with the interpretation vectors also converges faster than GBDT2NN model.

\begin{table}[!t]
\caption{Prediction results of the joint leaning method.}
\label{tbl:pred}
\centering
\begin{tabular}{l|ccccc}
\hline
Dataset & GBDT & GBDT2NN & Joint Method & $\lambda$ & metrics \\
\hline
AutoML-3 & 0.7634 & 0.7645 &0.7660 & 0.5 & AUC \\
\hline
OnlineNewsPop & 0.7387 & 0.7391 & 0.7398 & 0.7 & AUC \\
\hline
Zillow & 0.02197 & 0.02198 & 0.02187 & 0.7 & MSE\\
\hline
\end{tabular}
\end{table}

\section{Conclusion}
\label{sec:conclusion}

This paper proposes a novel explanation solution to explain GBDT2NN model.
It takes use of the interpreting vectors obtained from the structure-based tree explanation method to learn the features' attributions, and gives the explanation results based on the embedding.
In addition, we also put forward two concrete methods.
Empirical comparisons conducted on the benchmarks demonstrate the effectiveness of the proposed approaches on both interpretations and predictions, which also prove that imitating the explanation of the teacher model is a useful mean to improve the performance of distillation models.




\bibliographystyle{elsarticle-num}
\bibliography{main}





\end{document}